\documentclass[11pt, a4paper, onecolumn]{lumia}

\usepackage[sort&compress]{natbib}
\setcitestyle{authoryear,round,citesep={;},aysep={,},yysep={;}}

\usepackage{latexsym}
\usepackage{mathtools}
\usepackage{booktabs}
\usepackage{makecell}
\usepackage{tabularx}
\usepackage{longtable}
\usepackage{array}
\usepackage{url}
\usepackage{xurl}
\usepackage{graphicx}
\graphicspath{{figures/}}
\usepackage{float}
\usepackage{stfloats}
\definecolor{color5}{HTML}{006795}
\hypersetup{
  colorlinks = true,
  urlcolor   = color5,
  linkcolor  = color5,
  citecolor  = color5
}

\newcommand{\suggest}[1]{}

\setheadertext{LUMIA Lab}
\correspondingemail{\emailicon~boyizeng@sjtu.edu.cn, hyq718@sjtu.edu.cn, lin.zhouhan@gmail.com \quad $^*$ Equal Contribution \quad $^\ddagger$ Corresponding Author.}
\githublink{https://github.com/LUMIA-Group/Depth-Attention}
\huggingfacelink{https://huggingface.co/zeng123/Depth-attention-3B}
\renewcommand{\today}{2026-05-30}

\title{Depth-Attention: Cross-Layer Value Mixing for Language Models}

\author{%
    \textbf{Boyi Zeng}$^{1,*}$,
    \textbf{Yiqin Hao}$^{1,*}$,
    \textbf{Zitong Wang}$^{1,4}$,
    \textbf{Shixiang Song}$^{1,5}$,
    \textbf{He Li}$^{1}$,
    \textbf{Feichen Song}$^{1}$,
    \textbf{Yifan Liu}$^{1}$,
    \textbf{Ziwei He}$^{5}$,
    \textbf{Xinbing Wang}$^{3}$,
    \textbf{Zhouhan Lin}$^{1,2\ddagger}$\\
    \scriptsize
    $^1$ LUMIA Lab, School of Artificial Intelligence, Shanghai Jiao Tong University \\
    $^2$ Shanghai AI Laboratory \quad $^3$ Shanghai Jiao Tong University \quad $^4$ Sun Yat-sen University \quad $^5$ Shanghai Innovation Institute
}

\begin{document}

\begin{abstract}
Self-attention selects information freely across the sequence, but across depth, Transformers merely add each layer's output to the residual stream, so later layers cannot selectively reuse earlier-layer representations. Recent cross-layer methods improve this flow but operate on hidden states outside attention, adding state beyond the key-value cache at inference---a cost that becomes increasingly salient as modern LLMs compress the cache with grouped-query and multi-head latent attention. We introduce \textbf{Depth-Attention}, which performs this selection inside the attention module itself: before a layer attends over the sequence, its query attends \emph{along depth} over the keys of earlier layers at the same position and mixes their values into the value that self-attention then reads. Because Depth-Attention reuses the standard attention queries, keys, and value-cache slots, storing depth-mixed values in place of the original values, it adds no parameters and introduces no persistent inference state beyond the standard key-value cache---the same cache size as a vanilla decoder and less than hidden-state-based cross-layer methods. On Qwen3-style decoders at 1.5B and 3B parameters, Depth-Attention attains the lowest perplexity and the highest average downstream accuracy, improving over the vanilla Transformer by up to 2.3 accuracy points and surpassing strong cross-layer baselines (mHC, Attention Residuals, DenseFormer) in perplexity and average accuracy, while adding under 0.01\% extra arithmetic FLOPs and no additional persistent inference state. The gains hold from 360M to 3B parameters and extend to looped Transformers.

\end{abstract}
\maketitle
\section{Introduction}
Transformer language models rely on self-attention to dynamically select information across the sequence dimension~\citep{vaswani2017attention}. Across depth, however, information flows far less selectively: standard decoders propagate earlier-layer information only through residual~\citep{he2016deep}, collapsing all previous layers into a single hidden state, so deeper layers cannot selectively retrieve the representation of any individual earlier layer.

A growing line of work makes this depth-wise flow more explicit. DenseFormer~\cite{pagliardini2024denseformer} aggregates the outputs of preceding layers through learned depth-wise averaging, and Hyper-Connections~\cite{zhu2025hyper} replace the single residual path with multiple interacting streams that exchange information across depths; most recently, Attention Residuals~\cite{team2026attention} make depth-wise selection adaptive by replacing fixed residual accumulation with a softmax attention over preceding layer outputs. These mechanisms all operate on hidden states, however: they require access to earlier layers' hidden representations, which lie outside the key-value cache. At inference, a model must therefore maintain or recompute this extra state on top of its key-value cache---an overhead that grows in relative terms as modern LLMs aggressively shrink the cache with grouped-query attention (GQA)~\cite{ainslie2023gqa} or multi-head latent attention (MLA)~\cite{liu2024deepseek}. It raises a natural question: can a layer reuse the keys and values it already stores to carry information across depth, without adding any memory at inference?

We propose \textbf{Depth-Attention}, which performs this depth-wise selection inside the attention module itself. Before a layer mixes information across the sequence with self-attention, it uses the current-layer query to attend, along depth, over the keys of earlier layers at the same token position, and mixes their values into an updated value state---the same attention operation as self-attention, but applied across layers instead of across the sequence. The layer then runs standard causal self-attention on this depth-mixed value, leaving its queries, keys, and the causal mask unchanged. Because Depth Attention reuses the queries and keys of self-attention and stores its depth-mixed values in place of the originals in the V cache, it adds no parameters and its only persistent inference state is the standard key-value cache, the same size as a vanilla decoder's---in contrast to hidden-state methods, which carry additional state on top of the cache in the form of widened residual streams or retained earlier-layer representations.

Across Qwen3~\cite{yang2025qwen3} architecture at 1.5B and 3B parameters, Depth-Attention attains the best average downstream accuracy and the lowest perplexity against the vanilla Transformer and recent cross-layer methods such as manifold hyper-connections (mHC)~\cite{xie2025mhc} and Attention Residuals, while incurring lower compute and memory overhead than these baselines.

% We evaluate Depth-Attention on Qwen-style Transformer decoders at the 3B scale.
% Our method achieves the best performance among the compared architectures, demonstrating the effectiveness of adaptive depth-wise value mixing for language model pretraining.
% At the same time, Depth-Attention introduces only minor runtime overhead and requires no additional memory storage during inference.
% These results show that cross-layer information reuse can be made both effective and efficient by moving depth-wise selection from the hidden-state level into the attention module itself.
\section{Related Work}
\label{sec:related_work}

Residual connections~\cite{he2016deep} are the standard mechanism for passing information across layers in deep networks.
In Transformer decoders~\cite{vaswani2017attention}, each block updates the hidden state by adding its newly computed transformation to the representation inherited from the previous block.
This design greatly improves the optimization of deep models, and a large body of work has further studied how normalization placement~\citep{xiong2020layer}, residual scaling~\citep{wang2024deepnet}, gating~\citep{bachlechner2021rezero}, and skip-connection variants affect training stability~\citep{xie2023residual}.
However, the standard residual pathway also compresses the computation history into a single hidden state.
After the information produced by earlier layers has been merged into this state, later layers have no direct way to revisit a specific earlier representation.

A straightforward solution is to expose intermediate representations more explicitly.
DenseNet~\cite{huang2017densely} connects each layer to all preceding feature maps, allowing later layers to reuse earlier computations directly, while~\cite{peters2018deep,zhou2025value} learns scalar combinations of representations from different layers.
Transformer variants follow a similar intuition.
DenseFormer~\cite{pagliardini2024denseformer} and ANCRe~\cite{zhang2026ancre} introduce learned layer-wise coefficients to combine historical representations, and RealFormer~\cite{he2021realformer} carries attention scores across consecutive layers.
These methods suggest that intermediate-layer representations provide useful information beyond the final hidden state.
Nevertheless, many of them either use input-independent layer weights or propagate attention scores, rather than enabling each token to selectively choose the value information it reads from previous layers.

More recent architectures introduce more flexible cross-layer communication.
Hyper-Connections and its variants~\cite{zhu2025hyper,xie2025mhc} replace the single residual state with multiple interacting streams.
MUDDFormer~\cite{xiao2025muddformer} predicts position-dependent connection weights for separated query, key, value, and residual streams.
Other methods, including LAuReL~\cite{menghani2024laurel}, Dreamer~\cite{knupp2026depth}, and Attention Residuals~\cite{team2026attention}, design more specialized mechanisms for reusing earlier activations or layer outputs.
These approaches improve access to intermediate representations from different perspectives, but their cross-layer reuse is still mainly built on hidden representations.
When applied to deep autoregressive models, retaining or communicating such intermediate hidden states can introduce additional memory or communication overhead.

Depth-Attention follows the same general direction of making earlier-layer information more accessible, but places the cross-layer interaction at the value-state level.
Instead of storing an additional history of full hidden states, it reuses value states in a form compatible with the existing KV-cache structure.
This allows the model to mix information across layers while avoiding the extra memory cost associated with hidden-state-based reuse.
% 怎么还在写 depth-softmax，没有讲出在 value 相比于 hidden state 的优点,related work也不用方法写这么一大通
\begin{figure*}[t]
\centering
\includegraphics[width=\linewidth]{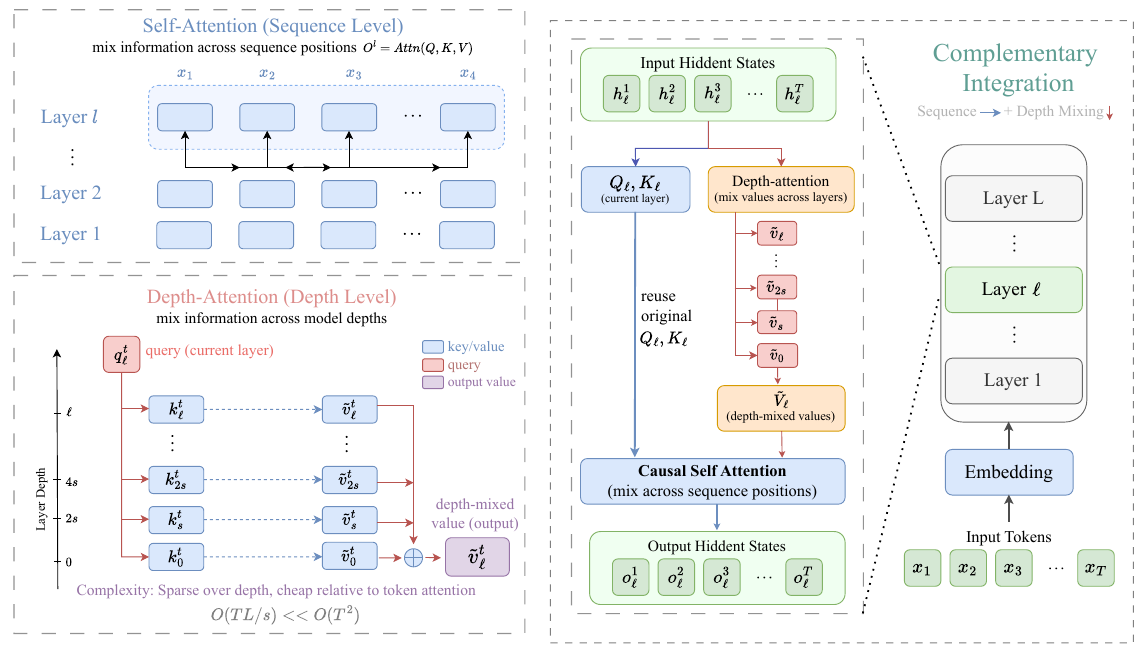}
\caption{Overview of \textbf{Depth-Attention}. Standard self-attention mixes information across sequence positions within each layer, whereas Depth-Attention performs a complementary operation across model depths at the same token position. For token position $t$ in layer $\ell$, the current-layer query $q_\ell^t$ attends to keys from the current layer and a sparse, strided set of shallower layers, e.g., $\{k_\ell^t, k_0^t, k_s^t, k_{2s}^t, \ldots\}$ restricted to layers below $\ell$, and uses the resulting depth-wise attention weights to combine the current value with previously depth-mixed values into $\widetilde{v}_\ell^t$. Because the depth source set is sparse, Depth-Attention adds only $O(TL/s)$ operations per layer, which is negligible compared with the $O(T^2)$ cost of token-level self-attention. Applying this operation to all token positions produces the depth-mixed value state $\widetilde{V}_\ell$, which replaces the original value state $V_\ell$ in the subsequent causal self-attention while keeping the queries, keys, and causal mask unchanged. Depth-Attention can be seamlessly integrated into Transformer layers with negligible overhead.}
\label{fig:main_fig}
\end{figure*}

\section{Methodology}
\label{sec:methodology}

\subsection{Overview}
Self-attention lets each token mix information across the sequence, but offers no comparable mechanism across depth. \textbf{Depth-Attention} fills this gap with an analogous operation along the depth dimension: at a fixed token position, the current-layer query attends over the keys of the current and earlier layers and mixes their values into a single depth-mixed value. A layer first applies Depth-Attention to update its value, then runs standard causal self-attention that reads from this depth-mixed value, leaving the queries, keys, and causal mask unchanged. 
Because Depth-Attention reuses the query and key already computed by self-attention, and stores the depth-mixed value in the same V-cache slot as the original, it adds no parameters and its only persistent inference state is the standard key-value cache, unchanged in size from a vanilla decoder's.

\subsection{Depth-Attention}

Consider a decoder layer $\ell$ operating on a sequence of length $T$, with query, key, and value states
\begin{equation}
Q_\ell, K_\ell, V_\ell \in \mathbb{R}^{T \times d},
\end{equation}
where $d$ is the head dimension; $q_\ell^t$, $k_\ell^t$, and $v_\ell^t$ denote the states of the token at position $t$.

Depth-Attention is a standard attention operation applied along the \emph{depth} dimension at a fixed token position.
For the token at position $t$ in layer $\ell$, the current query $q_\ell^t$ attends over the keys from the current and earlier layers at the same position:
\begin{equation}
\alpha_{\ell,j}^t
=
\operatorname{softmax}_{j \le \ell}
\left(
\frac{q_\ell^t \cdot k_j^t}{\sqrt{d}}
\right),
\quad j \le \ell .
\end{equation}

Using these depth-attention weights, Depth-Attention forms a depth-mixed value by reading from the current-layer value and the depth-mixed values of earlier layers:
\begin{equation}
\widetilde{v}_\ell^t
=
\alpha_{\ell,\ell}^t v_\ell^t
+
\sum_{j < \ell}
\alpha_{\ell,j}^t \widetilde{v}_j^t .
\end{equation}
The current layer contributes its own value $v_\ell^t$, while each earlier layer $j < \ell$ contributes its depth-mixed value $\widetilde{v}_j^t$, so that information accumulated at shallower depths is propagated upward.
Applying this to every position gives the depth-mixed value state $\widetilde{V}_\ell \in \mathbb{R}^{T \times d}$.

The layer then performs standard causal self-attention, reading from the depth-mixed value while keeping the original queries and keys:
\begin{equation}
O_\ell = \operatorname{CausalAttn}\!\left( Q_\ell, K_\ell, \widetilde{V}_\ell \right).
\end{equation}
Depth-Attention thus replaces only the value consumed by self-attention: the queries, keys, and causal mask remain those of a vanilla decoder.
At inference, layer $\ell$ stores its depth-mixed value $\widetilde{V}_\ell$ in the standard V-cache slot \emph{in place of} $V_\ell$; because the two have identical shape, this reuses the existing key-value cache without enlarging it. Subsequent tokens read $\widetilde{V}_\ell$ from that slot both as the value of layer $\ell$'s causal self-attention and, when needed, as the depth source for any later layer that mixes through it---so Depth-Attention introduces no persistent inference state beyond what a vanilla decoder already maintains.
Because the attention runs over the depth dimension, whose size (the number of layers) is far smaller than the sequence length $T$, its overhead is negligible compared with sequence-wise self-attention.
In practice, the index $j$ in the equations above ranges over a sparse strided subset of earlier layers rather than the full $j \le \ell$; the form of the operation is unchanged, only the number of source layers summed over, as detailed in the next subsection.
\subsection{Efficient Implementation}

\paragraph{Sparse depth sources via stride.}
Although the per-layer overhead of Depth-Attention is asymptotically negligible, allowing layer $\ell$ to attend to \emph{every} earlier layer makes the depth source set grow with $\ell$. This inflates activation memory and---under pipeline parallelism, where consecutive layers live on different devices---the cross-stage communication required to fetch earlier-layer keys and values.
To bound both costs, we expose each layer only to a strided subset of earlier layers: with zero-indexed layer ids and stride $s$, layer $\ell$ attends to its own value and to the values of layers $0, s, 2s, \ldots$ that are below $\ell$.
Equivalently, its depth source set is
\[
\mathcal{D}_{\ell}
=
\{\ell\}
\cup
\{\, ms \mid m \ge 0,\ ms < \ell \,\}.
\]
We find that this sparsification does not hurt model quality compared with denser source sets: our ablations (\autoref{fig:ablation_stride}) show that a stride of $s = L/2$, where $L$ is the total number of layers, gives the best results, and we use this setting as the default for all subsequent experiments unless otherwise specified.

\paragraph{Grouped-query attention.}
Modern LLMs widely adopt grouped-query attention (GQA), which shrinks the key-value cache by letting $g$ query heads share a single key-value head.
Depth-Attention adapts naturally to this setting: we average each group of $g$ queries into a single query so that the depth-wise attention runs at the key-value head resolution.
Because the key-value dimension is several times smaller than the hidden size, performing Depth-Attention at this resolution further shrinks both its extra compute and its extra memory, making our method especially efficient in this widely-used regime.

\section{Experiments}
\label{sec:experiments}

Our experiments are organized into five parts.
First, we conduct large-scale pretraining experiments on Qwen3-style architectures at 1.5B and 3B scales, comparing Depth-Attention with vanilla Transformer decoders and representative cross-layer baselines.
Second, we analyze the efficiency of Depth-Attention from both theoretical and empirical perspectives, covering computation, training wall-clock overhead, inference throughput, and memory usage.
Third, we study the scaling behavior of Depth-Attention across model sizes, evaluating whether its gains remain consistent on different model scales.
Fourth, we visualize the learned depth-attention weights to examine how the model uses shallower value sources when constructing depth-mixed value states.
Finally, we evaluate Depth-Attention in a looped Transformer setting, testing whether the same depth-wise value mixing mechanism can extend to recurrent depth induced by parameter sharing.
\suggest{我喜欢类似下面这种写法：Our experiments consist of 3 parts. First, we validate the scaling curves of pondering models on widely
used GPT-2 and LLaMA architectures. Second, we perform large-scale pretraining of PonderPythia
models on the Pile dataset and compare their scaling curves and language modeling capabilities with
those of the official Pythia suite (Biderman et al., 2023). Third, we evaluate the downstream task
performance of PonderPythia models, including 9 popular general tasks and an instruction-following
task, and compare the results with official Pythia, OPT (Zhang et al., 2022), Bloom (Le Scao et al.,
2023) and TinyLLaMA (Zhang et al., 2024).}

\subsection{Main Results}

We first evaluate Depth-Attention on Qwen3-style decoder architectures~\cite{yang2025qwen3} at 1.5B and 3B scales.
All models are trained from scratch for 32B tokens on the Pile~\cite{gao2020pile} under the same pretraining protocol, including the same training data, optimization schedule, and hyperparameters.
We compare Depth-Attention with the vanilla Transformer decoder and representative cross-layer baselines, including manifold hyperconnection (mHC)~\cite{xie2025mhc}, Attention Residuals~\cite{team2026attention}, and DenseFormer-style dense layer reuse~\cite{pagliardini2024denseformer}.

We report Pile validation perplexity to measure language modeling quality, and evaluate downstream performance under both zero-shot and five-shot settings, including LAMBADA~\cite{paperno2016lambada}, PIQA~\cite{bisk2020piqa}, WinoGrande~\cite{sakaguchi2021winogrande}, SciQ~\cite{welbl2017crowdsourcing}, HellaSwag~\cite{zellers2019hellaswag}, ARC-Easy, ARC-Challenge~\cite{clark2018think}, and RACE~\cite{lai2017race}.
Detailed model configurations and training hyperparameters are provided in~\autoref{app:training-details}.

\begin{table*}[t]
\centering
\small
\setlength{\tabcolsep}{1.6pt}
\renewcommand{\arraystretch}{1.06}
\begin{tabular*}{\textwidth}{@{\extracolsep{\fill}}llcccccccccc@{}}
\toprule
\textbf{Size} &
\textbf{Method} &
\textbf{PPL} $\downarrow$ &
\textbf{LAMBADA} &
\textbf{PIQA} &
\shortstack[c]{\textbf{Wino}\\\textbf{Grande}} &
\textbf{SciQ} &
\shortstack[c]{\textbf{Hella}\\\textbf{Swag}} &
\shortstack[c]{\textbf{ARC-}\\\textbf{Easy}} &
\shortstack[c]{\textbf{ARC-}\\\textbf{Challenge}} &
\textbf{RACE} &
\textbf{Avg.} \\
\midrule
\multicolumn{12}{c}{\textbf{Zero-shot evaluation}} \\
\midrule
1.5B & Vanilla
& 8.17 & 55.02 & 68.61 & 54.85 & 84.10 & 36.57 & 54.08 & 23.72 & 33.11 & 51.26 \\
1.5B & mHC
& 7.88 & 57.77 & 69.10 & \textbf{56.75} & 86.60 & 37.88 & 56.52 & 24.83 & 33.59 & 52.88 \\
1.5B & Attention Residuals
& 7.78 & 58.49 & 69.86 & 56.04 & 85.90 & 38.80 & 57.07 & 24.15 & 32.06 & 52.80 \\
1.5B & DenseFormer
& 7.88 & 56.34 & 69.26 & 56.27 & 85.40 & 38.47 & 56.36 & \textbf{25.43} & \textbf{33.78} & 52.66 \\
1.5B & \textbf{Depth-Attention}
& \textbf{7.77} & \textbf{59.89} & \textbf{70.24} & 55.72 & \textbf{88.20} & \textbf{38.86} & \textbf{58.50} & 24.49 & 32.92 & \textbf{53.60} \\
\midrule
3B & Vanilla
& 7.55 & 58.86 & 70.57 & 55.96 & 86.60 & 39.84 & 59.13 & 25.34 & 33.78 & 53.76 \\
3B & mHC
& 7.30 & 61.73 & 70.89 & 56.75 & 85.60 & 41.11 & 60.06 & 27.30 & 33.97 & 54.68 \\
3B & Attention Residuals
& 7.27 & 61.60 & 71.82 & 56.99 & 86.70 & 41.32 & \textbf{60.52} & \textbf{27.65} & 33.49 & 55.01 \\
3B & DenseFormer
& 7.35 & 59.91 & 71.82 & 58.17 & 86.90 & 40.95 & 59.43 & 26.11 & 34.35 & 54.70 \\
3B & \textbf{Depth-Attention}
& \textbf{7.25} & \textbf{62.25} & \textbf{72.36} & \textbf{60.38} & \textbf{88.00} & \textbf{41.88} & 60.14 & 25.51 & \textbf{35.02} & \textbf{55.69} \\
\midrule
\multicolumn{12}{c}{\textbf{Five-shot evaluation}} \\
\midrule
1.5B & Vanilla
& -- & 48.71 & 69.48 & 53.28 & 90.10 & 36.60 & 57.95 & 25.09 & 31.96 & 51.65 \\
1.5B & mHC
& -- & 51.06 & 70.08 & \textbf{56.67} & 91.80 & 38.24 & 59.30 & \textbf{27.30} & 33.01 & 53.43 \\
1.5B & Attention Residuals
& -- & 53.13 & 70.08 & 56.43 & \textbf{91.90} & 38.55 & \textbf{60.19} & 26.02 & 33.01 & 53.66 \\
1.5B & DenseFormer
& -- & 52.71 & \textbf{71.16} & \textbf{56.67} & 91.00 & 38.37 & 58.46 & 26.45 & \textbf{34.45} & 53.66 \\
1.5B & \textbf{Depth-Attention}
& -- & \textbf{54.05} & 70.51 & 55.56 & 91.50 & \textbf{38.59} & 59.55 & 25.94 & 34.07 & \textbf{53.72} \\
\midrule
3B & Vanilla
& -- & 53.23 & 71.16 & 55.01 & 91.20 & 39.79 & 61.53 & 27.90 & 33.49 & 54.16 \\
3B & mHC
& -- & 56.76 & 71.87 & 58.64 & \textbf{93.50} & 40.86 & 62.58 & 27.22 & 33.78 & 55.65 \\
3B & Attention Residuals
& -- & 55.73 & 72.63 & 57.46 & 92.70 & 41.07 & \textbf{64.31} & \textbf{28.67} & 34.83 & 55.92 \\
3B & DenseFormer
& -- & 54.12 & 72.47 & \textbf{59.27} & 91.90 & 40.71 & 63.80 & 27.99 & 34.45 & 55.59 \\
3B & \textbf{Depth-Attention}
& -- & \textbf{57.40} & \textbf{72.91} & 59.19 & 92.80 & \textbf{41.51} & 63.22 & 26.71 & \textbf{35.60} & \textbf{56.17} \\
\bottomrule
\end{tabular*}
\caption{
Main results on Qwen3-style architectures.
We report validation perplexity and downstream accuracy under zero-shot and five-shot evaluation settings.
Since perplexity is independent of the shot setting, we report it only in the zero-shot block and use ``--'' in the five-shot block.
All downstream numbers are accuracies in percentage.
Avg. is averaged over the eight downstream tasks shown in the table.
}
\label{tab:qwen-main-results}
\end{table*}

As shown in~\autoref{tab:qwen-main-results}, Depth-Attention attains the best perplexity and the best average downstream accuracy against the vanilla Transformer and all cross-layer baselines on both 1.5B and 3B Qwen3-style models, under both zero-shot and five-shot evaluation; per-task trends are mixed, but no individual baseline matches Depth-Attention's average at either scale or evaluation setting.
These results demonstrate that Depth-Attention effectively improves model performance by enriching current-layer value states through depth-wise value mixing.

\suggest{这部分只有 qwen，大表像 latent cot 那样给细分任务，不要直给平均值，位置不够的话可以去掉Lambada
(Std)} 
\suggest{dynamically routing这种写法不太对吧，别人 还以为是加个 router，后面也是} 
\subsection{Efficiency Analysis}
\label{sec:efficiency}
\paragraph{Theoretical overhead.}
We compare three sources of overhead that cross-layer mechanisms add to a vanilla decoder: extra computation (FLOPs), extra activation I/O, and extra memory for long contexts.
\autoref{tab:theoretical-overhead} reports the per-token extra FLOPs and I/O at a target layer $\ell$, instantiates their totals for one token under our 3B Qwen3-style configuration, and lists the theoretical extra memory required to prefill a 128K-token context.
Depth-Attention adds far less computation and activation I/O than representative cross-layer baselines.
Moreover, because it reuses the standard key-value cache, Depth-Attention adds no persistent state at inference: its extra prefill memory is zero, whereas the hidden-state baselines require gigabytes of additional memory at long context (up to $24$~GiB for DenseFormer at 128K tokens).
Detailed derivations are provided in~\autoref{app:overhead-details}.

\begin{table*}[t]
\centering
\small
\setlength{\tabcolsep}{3pt}
\renewcommand{\arraystretch}{1.12}
\begin{tabular*}{\textwidth}{@{\extracolsep{\fill}}lccccc@{}}
\toprule
\textbf{Method}
& \shortstack{\textbf{Layer-$\ell$}\\\textbf{Extra FLOPs}}
& \shortstack{\textbf{3B Extra}\\\textbf{FLOPs / Token}}
& \shortstack{\textbf{Layer-$\ell$}\\\textbf{Extra I/O}}
& \shortstack{\textbf{3B Extra}\\\textbf{I/O / Token}}
& \shortstack{\textbf{128K Prefill}\\\textbf{Extra Memory}} \\
\midrule
\textbf{Depth-Attention}
& $O\!\left((\lceil \ell/s\rceil+1)d_{\mathrm{kv}}\right)$
& \textbf{$\mathbf{2.42{\times}10^{5}}$}
& $O\!\left((\lceil \ell/s\rceil+1)d_{\mathrm{kv}}\right)$
& \textbf{$\mathbf{284}$ KiB}
& $\mathbf{0}$ \textbf{GiB} \\
DenseFormer
& $O(\ell d_{\mathrm{model}})$
& $5.01{\times}10^{6}$
& $O(\ell d_{\mathrm{model}})$
& $4.97$ MiB
& $24.0$ GiB \\
Attention Residuals
& $O(R_\ell d_{\mathrm{model}})$
& $4.26{\times}10^{6}$
& $O(R_\ell d_{\mathrm{model}})$
& $4.44$ MiB
& $4.0$ GiB \\
mHC
& $O(N^3d_{\mathrm{model}})$
& $4.72{\times}10^{7}$
& $O(Nd_{\mathrm{model}})$
& $8.63$ MiB
& $1.5$ GiB \\
\bottomrule
\end{tabular*}
\caption{
Theoretical comparison of additional computation, activation I/O, and long-context memory beyond the vanilla Transformer decoder.
The FLOPs and I/O formulas report the extra per-token overhead at target layer $\ell$, while the 3B columns instantiate the total overhead for one token over the full model.
Extra I/O denotes additional activation read/write over the vanilla decoder; for methods that replace the standard residual connection, we subtract the vanilla residual I/O.
The 128K-prefill memory column reports the extra persistent memory required to prefill a 128K-token context.
The estimates use our 3B Qwen3-style configuration with $L=48$, $d_{\mathrm{model}}=2048$, $d_{\mathrm{kv}}=512$, and Depth-Attention stride $s=24$.
Here $R_\ell$ is the number of residual blocks used by Attention Residuals at layer $\ell$, and $N$ is the number of mHC streams.
Detailed derivations are provided in Appendix~\ref{app:overhead-details}.
}
\label{tab:theoretical-overhead}
\end{table*}

\paragraph{Training wall-clock.}
Since none of the cross-layer baselines provides an official kernel-fused implementation, we benchmark every method---including Depth-Attention---under a common reference PyTorch implementation, with no custom kernels.
This setting penalizes all methods relative to a fully optimized stack, but it isolates the \emph{methodological} overhead and puts every baseline on the same footing as our method.
\autoref{tab:wallclock} reports the per-step training time for the 1.5B and 3B Qwen3-style configurations used in our main experiments.
Depth-Attention adds only $+9.1\%$ and $+11.2\%$ wall-clock at the two scales---the smallest overhead of any cross-layer mechanism by a wide margin: DenseFormer and Attention Residuals are roughly $4\times$--$7\times$ further from vanilla than Depth-Attention, and mHC is more than $25\times$ further.
A dedicated fused kernel, which we leave to future work, would likely close even this small gap.

\par\medskip
\noindent
\begin{minipage}[t]{0.50\textwidth}
\vspace{0pt}
\centering
\footnotesize
\setlength{\tabcolsep}{1.8pt}
\renewcommand{\arraystretch}{1.16}
\begin{tabularx}{\linewidth}{@{}>{\raggedright\arraybackslash}Xcc@{}}
\toprule
\textbf{Method} & \textbf{1.5B} & \textbf{3B} \\
\midrule
Vanilla                  & $3.28$ & $4.63$ \\
\textbf{Depth-Attention} & $\mathbf{3.58}~(\mathbf{+9.1\%})$ & $\mathbf{5.15}~(\mathbf{+11.2\%})$ \\
DenseFormer              & $4.95~(+50.6\%)$ & $6.84~(+47.9\%)$ \\
\makecell[l]{Attention\\Residuals}      & $4.85~(+47.8\%)$ & $8.10~(+75.1\%)^{\dagger}$ \\
mHC                      & $13.83~(+321\%)$ & $18.85~(+308\%)$ \\
\bottomrule
\end{tabularx}
\end{minipage}\hfill
\begin{minipage}[t]{0.46\textwidth}
\vspace{0pt}
\captionsetup{justification=raggedright,singlelinecheck=false}
\captionof{table}{
Per-step training wall-clock on Qwen3-style 1.5B and 3B configurations.
Each cell reports sec/step and overhead over vanilla using the same reference PyTorch implementation.
$^{\dagger}$Attention Residuals at 3B uses bs=2, ga=16 because the shared setting does not fit.
}
\label{tab:wallclock}
\vspace{0pt}
\end{minipage}
\par\medskip

\paragraph{Inference efficiency.}
We further measure the empirical inference cost of Depth-Attention on the trained 3B Qwen3-style model.
With batch size 64, prefill length 2048, and decode length 2048, total generation time rises from $526.6$\,s to $532.8$\,s---an overhead of only $+1.18\%$---while peak GPU memory during prefill is unchanged at $30.6$\,GB.
Combined with the negligible extra arithmetic cost ($+0.004\%$ FLOPs, normalized by the standard dense one-token decoding estimate $2N$~\citep{kaplan2020scaling}), Depth-Attention adds essentially no latency and no additional persistent inference state, consistent with the structural fact that it stores its depth-mixed values in the standard V-cache slot rather than in any new buffer or wider residual stream.

\par\medskip
\noindent
\begin{minipage}[t]{0.42\textwidth}
\vspace{0pt}
\centering
\scriptsize
\setlength{\tabcolsep}{1.5pt}
\renewcommand{\arraystretch}{1.45}
\begin{tabularx}{\linewidth}{@{}>{\raggedright\arraybackslash}Xccc@{}}
\toprule
\textbf{Method}
& \shortstack[c]{\textbf{Extra}\\\textbf{FLOPs}}
& \shortstack[c]{\textbf{Inference}\\\textbf{Time (s)}}
& \shortstack[c]{\textbf{Prefill}\\\textbf{Memory (GB)}} \\
\midrule
Vanilla
& 0
& 526.6
& 30.6 \\
Depth-Attention
& $2.42{\times}10^{5}$
& 532.8
& 30.6 \\
\midrule
Overhead
& $+0.004\%$
& $+1.18\%$
& $+0.000$ \\
\bottomrule
\end{tabularx}
\end{minipage}\hfill
\begin{minipage}[t]{0.54\textwidth}
\vspace{0pt}
\captionsetup{justification=raggedright,singlelinecheck=false}
\captionof{table}{
Inference efficiency on the 3B Qwen3-style model.
Time is measured with batch size 64, prefill length 2048, and decode length 2048.
Extra FLOPs reports the additional depth-wise attention cost per token, normalized by $2N$.
}
\label{tab:efficiency}
\vspace{0pt}
\end{minipage}
\par\medskip
\Needspace{0.48\textheight}
\subsection{Scaling Behavior}
\label{sec:scaling}

We further evaluate whether the benefit of Depth-Attention persists across model scales.
To this end, we train Qwen-style decoder architectures at multiple model sizes under the same pretraining setting, and compare Depth-Attention with the corresponding vanilla Transformer baseline.
Detailed model configurations and optimization hyperparameters are provided in~\autoref{app:scaling-details}.

As shown in~\autoref{fig:scaling_law}, Depth-Attention consistently achieves lower validation loss than vanilla Transformers across all tested model sizes.
This improvement remains stable from 360M to 3B parameters, indicating that the benefit of depth-wise value mixing persists as the model scales up.
These results suggest that Depth-Attention improves the scaling behavior of Transformer decoders by enriching current-layer value states with information from earlier depths.
\suggest{后面加上 llama 的结果，说明我们的方法跨架构有效，无论是 qwen 还是 llama，拟合具体倍数，简洁一点，参考ponder1 论文的写法}

\par\medskip
\noindent
\begin{minipage}[t]{0.48\textwidth}
    \vspace{0pt}
    \centering
    \includegraphics[width=\linewidth]{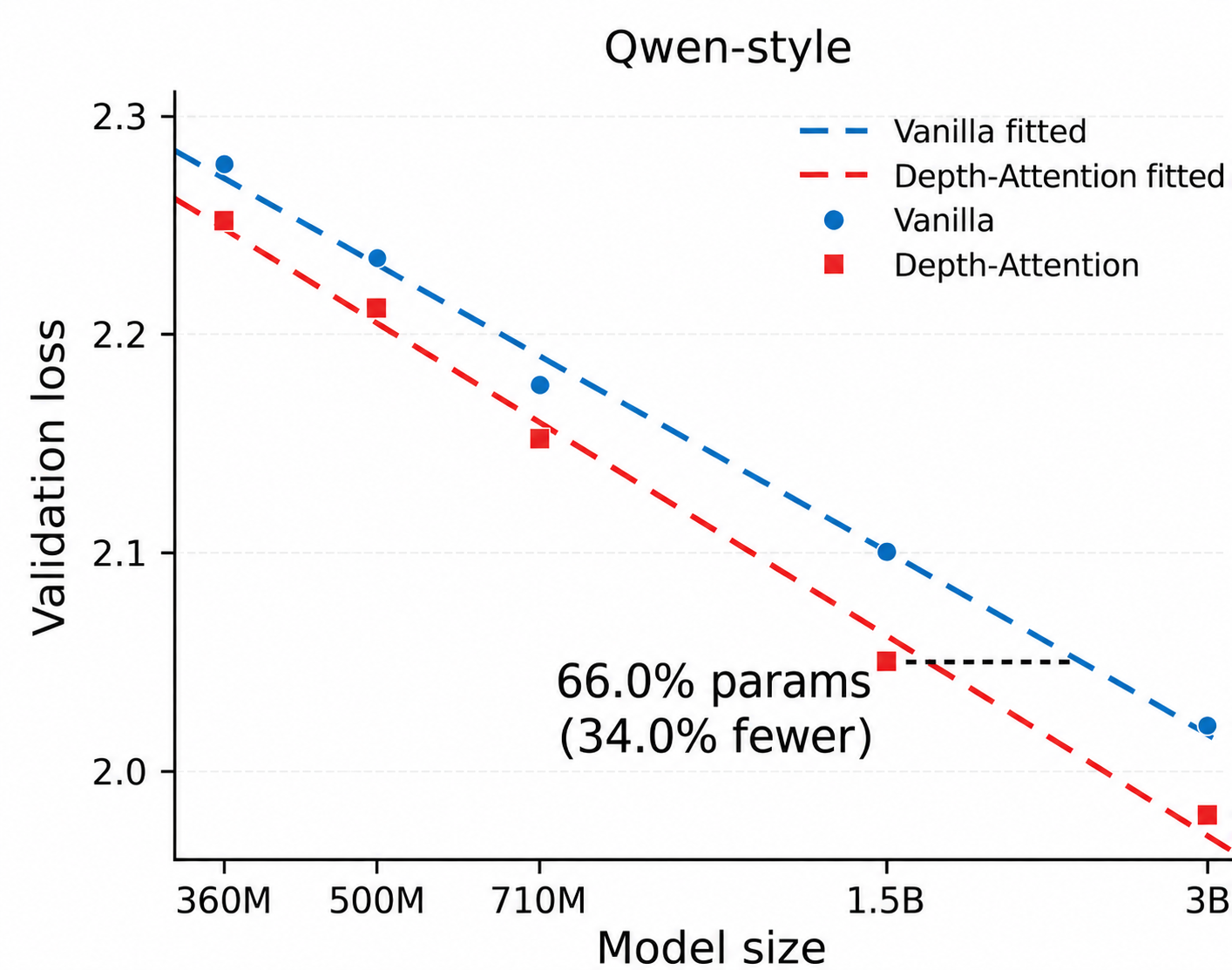}
\end{minipage}\hfill
\begin{minipage}[t]{0.48\textwidth}
    \vspace{0pt}
    \captionsetup{justification=raggedright,singlelinecheck=false}
    \captionof{figure}{
    Scaling behavior on Qwen-style architectures under the same pretraining setting.
    Depth-Attention consistently shifts the validation-loss curve downward from 360M to 3B parameters, showing that depth-wise value mixing remains useful as model size increases.
    Dashed curves denote fitted scaling trends, and the black dashed segment indicates the estimated parameter reduction needed for a vanilla baseline to match the 1.5B Depth-Attention model.
    }
    \label{fig:scaling_law}
    \vspace{0pt}
\end{minipage}

\subsection{Depth-Attention in Looped Transformers}

Looped, or recurrent, Transformers have recently attracted renewed interest as a way to increase effective depth through parameter sharing rather than by adding new layers~\citep{giannou2023looped,zeng2026ponderlm,bae2026mixture}.
% TODO: cite Ponder / looped-transformer (latent-reasoning) works in the sentence above
Depth-Attention applies directly to this setting: when a layer is executed repeatedly, the value states it produced at earlier loop steps serve as additional depth sources, which the current step mixes into its value before standard causal self-attention.
On a 500M looped Transformer with three loops, adding Depth-Attention lowers the validation loss from $2.208$ to $2.194$, showing that depth-wise value mixing remains effective when depth is induced by recurrence rather than by stacking layers.

\subsection{Analysis of Depth-Attention Weights}
To examine whether Depth-Attention actually uses shallower value sources, we visualize the learned depth-attention weights of the trained 3B model in~\autoref{fig:depth_weight_heatmap}.
To quantify how each target layer uses sources in its depth source set, we calculate the average depth-attention weight for every target-source layer pair over tokens, heads, and evaluation samples.

As shown in~\autoref{fig:depth_weight_heatmap}, the learned weights are not concentrated solely on the current-layer source.

\par\medskip
\noindent
\begin{minipage}[t]{0.48\textwidth}
\vspace{0pt}
\centering
\includegraphics[width=0.82\linewidth]{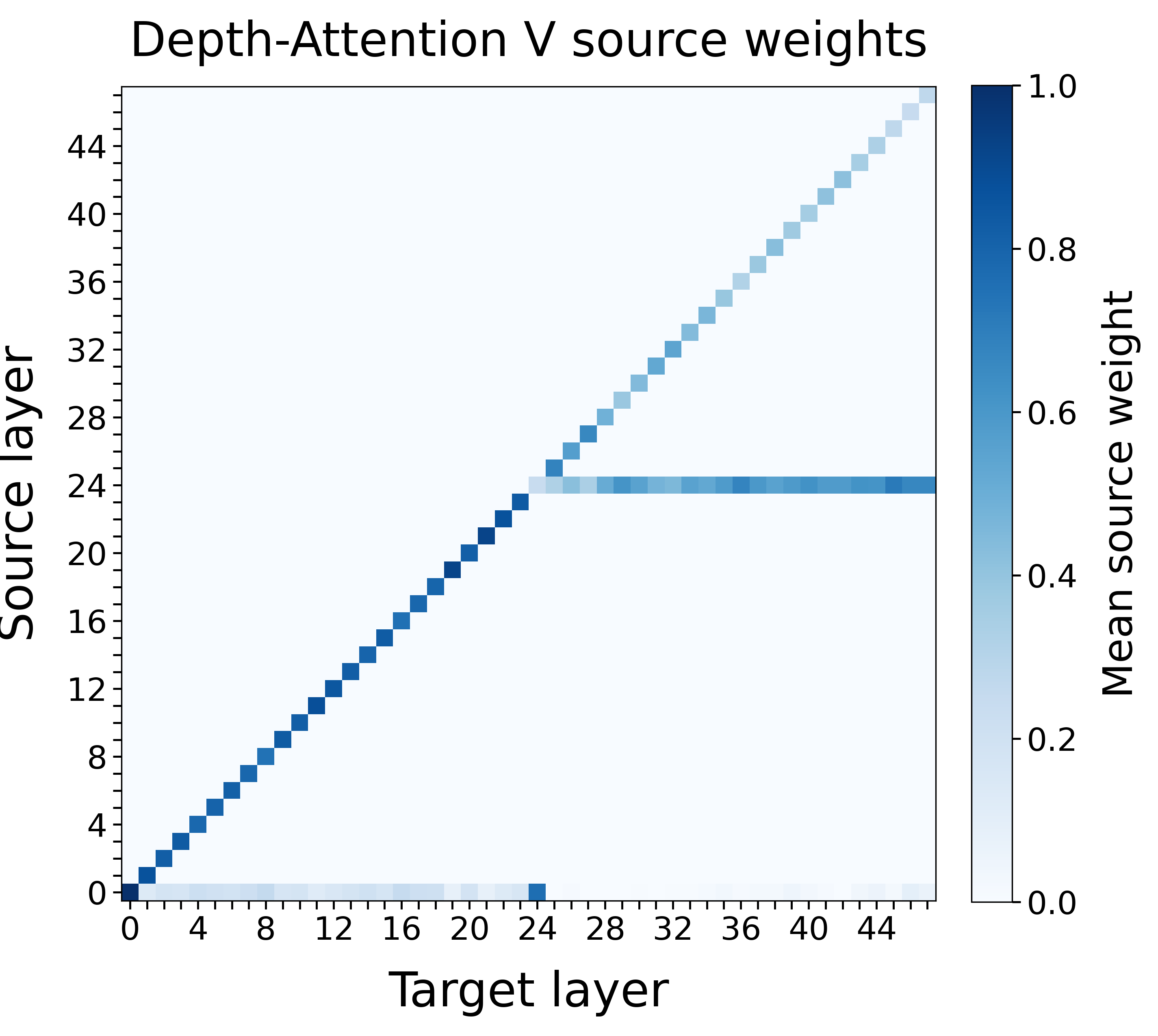}
\end{minipage}\hfill
\begin{minipage}[t]{0.48\textwidth}
\vspace{0pt}
\captionsetup{justification=raggedright,singlelinecheck=false}
\captionof{figure}{
Visualization of learned depth-attention weights in the trained 3B model, averaged over tokens, heads, and evaluation samples.
The diagonal corresponds to the current-layer value source, while off-diagonal entries show the reuse of shallower value states.
The visible mass on earlier layers indicates that Depth-Attention does not collapse to vanilla attention, but learns to mix information from multiple depths.
}
\label{fig:depth_weight_heatmap}
\vspace{0pt}
\end{minipage}

Besides the diagonal pattern, deeper target layers assign clear weight to shallower source layers, with a particularly visible band around the intermediate layers.
This shows that Depth-Attention does not collapse to the vanilla current-layer computation; instead, it learns to enrich the current-layer value state with information from earlier depths.
\suggest{Routing Analysis和前面一样的问题，router 不太对，换成最大的 3b 的}

\section{Ablation Studies}

We conduct controlled ablation studies using the same 500M Qwen-style configuration and training setup as in~\autoref{sec:scaling}. 
All other settings are kept unchanged, allowing us to isolate which design choices of Depth-Attention contribute to the observed improvement.
Specifically, we study three factors: the construction of the depth source list, the depth-mixing rule, and the attention states updated by depth mixing.

\paragraph{Stride.}
We first study how the density of the depth source list affects performance.
Specifically, we compare four Depth-Attention variants: using only the first-layer source, the default half-depth stride, a denser quarter-depth stride, and the full-source setting with stride $1$.
This ablation tests whether exposing each layer to more historical value states leads to consistent gains, and whether a sparse source list is sufficient to capture most of the benefit.

As shown in~\autoref{fig:ablation_stride}, the half-depth stride achieves the lowest validation loss.
Using only the first-layer source performs slightly worse, suggesting that an overly sparse source list does not provide enough useful intermediate-depth information.
However, making the source list denser does not further improve performance: both the quarter-depth and full-source settings yield slightly higher validation loss than the half-depth setting.
This suggests that overly dense source lists may introduce redundant or less useful historical value states.
Overall, the results indicate that both too few and too many depth sources are suboptimal, while a moderate source density works best.
For this reason, we use the half-depth stride as the default setting.
All four depth-mixing variants still outperform vanilla attention (validation loss $2.2348$), indicating that the benefit of depth-wise value reuse is robust across different source-list designs.
\par\medskip
\noindent
\begin{minipage}[t]{0.48\textwidth}
    \vspace{0pt}
    \centering
    \includegraphics[width=\linewidth]{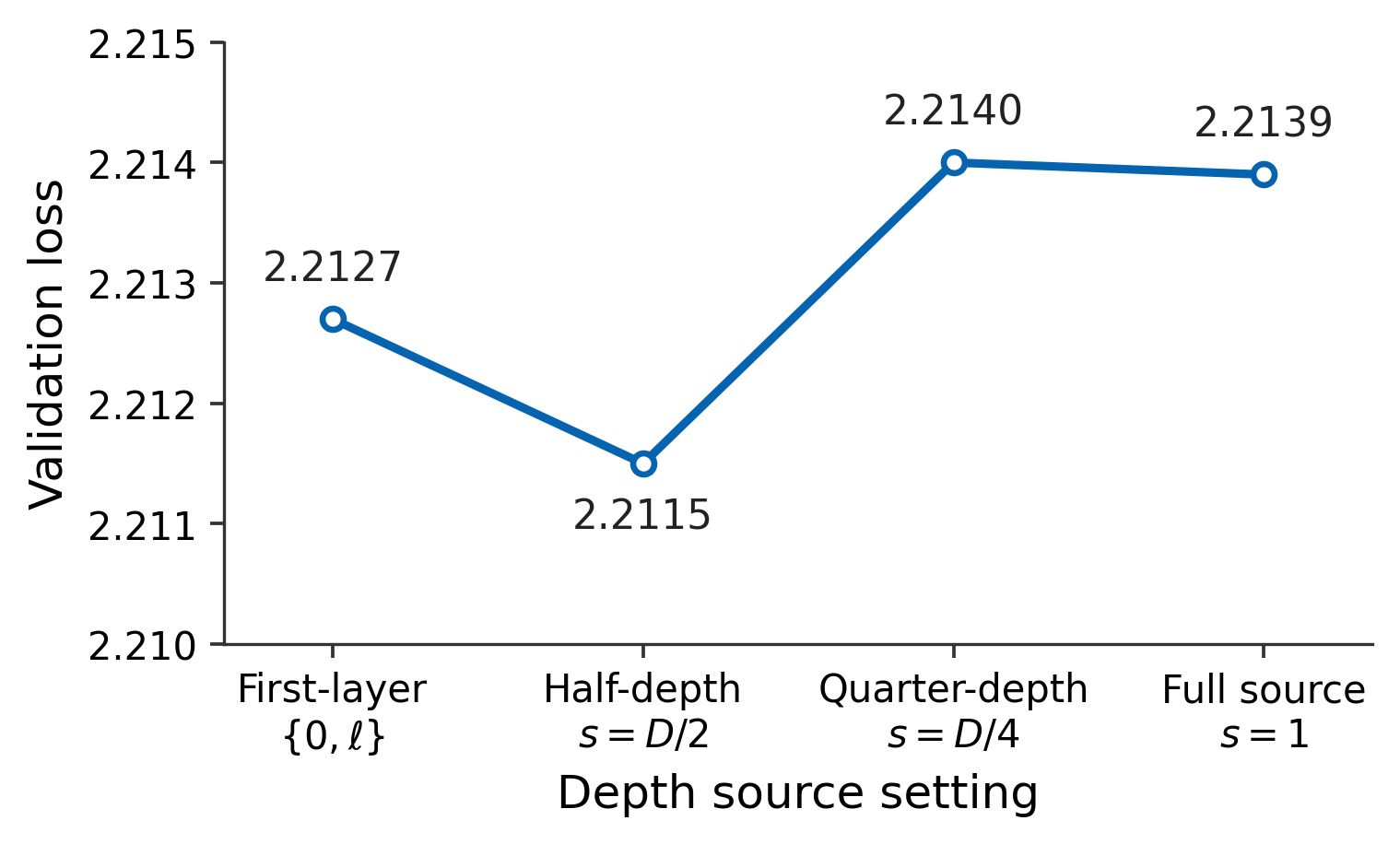}
\end{minipage}\hfill
\begin{minipage}[t]{0.48\textwidth}
    \vspace{0pt}
    \captionsetup{justification=raggedright,singlelinecheck=false}
    \captionof{figure}{
    Ablation on the stride used to construct the depth source list in the 500M model.
    The half-depth stride achieves the lowest validation loss among the first-layer-only, half-depth, quarter-depth, and full-source variants.
    This suggests that a moderate source density captures useful intermediate-depth information, while overly sparse or overly dense source lists are slightly less effective.
    }
    \label{fig:ablation_stride}
    \vspace{0pt}
\end{minipage}

\paragraph{Depth-mixing rule.}
We next ablate the rule used to mix value states across the depth source list.
Keeping the same source-list configuration, we compare vanilla attention, a uniform-mixing variant that simply averages all source values, and Depth-Attention, which computes per-head softmax weights over the depth source list.
This ablation examines whether adaptive depth-wise value mixing is more effective than merely exposing previous value states with fixed uniform weights.
As shown in~\autoref{tab:ablation_mixing_rule}, uniform mixing already improves over the vanilla baseline, indicating that reusing previous value states is beneficial.
Depth-Attention further reduces the validation loss, showing that per-head softmax depth mixing is more effective than a fixed average over the same source list.

\par\medskip
\noindent
\begin{minipage}[t]{0.56\textwidth}
\vspace{0pt}
\centering
\small
\setlength{\tabcolsep}{4pt}
\renewcommand{\arraystretch}{1.08}
\begin{tabularx}{\linewidth}{@{}>{\raggedright\arraybackslash}X>{\raggedright\arraybackslash}Xc@{}}
\toprule
\textbf{Method} & \textbf{Mixing rule} & \textbf{Val. loss} \\
\midrule
Vanilla & -- & 2.2348 \\
Uniform Mix & Uniform weights & 2.2153 \\
Depth-Attention & Per-head softmax mixing & \textbf{2.2115} \\
\bottomrule
\end{tabularx}
\end{minipage}\hfill
\begin{minipage}[t]{0.40\textwidth}
\vspace{0pt}
\captionsetup{justification=raggedright,singlelinecheck=false}
\captionof{table}{
Ablation on the depth-mixing rule using the 500M model.
Uniform Mix and Depth-Attention use the same depth source list, while differing only in how source values are mixed.
}
\label{tab:ablation_mixing_rule}
\vspace{0pt}
\end{minipage}
\par\medskip

\paragraph{Depth-mixing object.}
Finally, we ablate which attention states should be updated by depth mixing.
Besides the proposed value-only update, we compare key-only and key-value variants, where the same depth-routing weights are used to construct mixed key states.
This ablation tests whether Depth-Attention should modify the keys used for temporal attention, or instead preserve the current-layer keys while reusing historical value states. As shown in~\autoref{tab:ablation_mixing_object}, value-only mixing achieves the best validation loss, showing that using values for depth mixing is beneficial. In contrast, mixing keys either hurts performance or weakens the gain, suggesting that Depth-Attention should preserve current-layer keys while updating only values.
\par\medskip
\noindent
\begin{minipage}[t]{0.56\textwidth}
\vspace{0pt}
\centering
\small
\setlength{\tabcolsep}{4pt}
\renewcommand{\arraystretch}{1.08}
\begin{tabularx}{\linewidth}{@{}>{\raggedright\arraybackslash}X>{\raggedright\arraybackslash}Xc@{}}
\toprule
\textbf{Method} & \textbf{Updated state} & \textbf{Val. loss} \\
\midrule
Vanilla & -- & 2.2348 \\
Key-only & $K_{\mathrm{mix}}$ & 2.2354 \\
Key-value & $K_{\mathrm{mix}}, V_{\mathrm{mix}}$ & 2.2198 \\
Value-only & $V_{\mathrm{mix}}$ & \textbf{2.2115} \\

\bottomrule
\end{tabularx}
\end{minipage}\hfill
\begin{minipage}[t]{0.40\textwidth}
\vspace{0pt}
\captionsetup{justification=raggedright,singlelinecheck=false}
\captionof{table}{
Ablation on the attention state updated by depth mixing using the 500M model.
All variants use the same depth-routing weights, but apply them to different attention states.
}
\label{tab:ablation_mixing_object}
\vspace{0pt}
\end{minipage}
\par\medskip
\section{Conclusion}

We introduced Depth-Attention, a simple way to bring adaptive, content-based selection to the depth dimension of Transformer decoders. Instead of modifying the residual stream, each layer's query attends along depth over the keys of earlier layers and mixes their values into the value state that self-attention reads, reusing the queries, keys, and values the model already computes. This makes cross-layer reuse attention-native: it adds no parameters, keeps standard key-value cache as its only persistent inference state (the same size as a vanilla decoder's, in contrast to methods that widen the residual stream or retain earlier-layer hidden states), and leaves the causal attention pattern unchanged. On Qwen3-style decoders at 1.5B and 3B parameters, Depth-Attention attains the best results over the vanilla Transformer and strong cross-layer baselines---at a small fraction of their compute and memory cost; the benefits hold across model scales and extend to looped Transformers.

\section*{Limitations}
Our study has several limitations. First, although Depth-Attention adds negligible FLOPs and no persistent state beyond the key-value cache in theory, our current implementation is not fused into the attention kernel, so the measured training and inference overhead is larger than the asymptotic analysis suggests; a dedicated kernel would likely recover most of this gap. Second, we validate Depth-Attention only up to 3B parameters and 32B training tokens on the Pile, so whether the gains persist at substantially larger model and data scales remains to be confirmed; this is primarily due to our limited computational resources rather than any inherent constraint of the method. Finally, we mix only value states and select depth sources with a fixed stride; learned source selection and richer cross-layer interactions remain open, although our ablations indicate that value-only mixing is the most effective variant we examined.

% \section*{Ethical Considerations}

% This work proposes an architectural modification to improve the performance and efficiency of Transformer language models. The method does not involve human-subject studies, private user data, or the collection of new datasets, and it is not designed for sensitive applications such as surveillance, profiling, or decision making about individuals. Since Depth-Attention only changes how intermediate value states are reused across layers, it does not introduce additional ethical risks beyond those already associated with standard language models. As with any language model, systems built with this method should still be evaluated for potential biases, harmful outputs, and misuse before deployment.

\bibliographystyle{acl_natbib}
\bibliography{custom}

\clearpage
\appendix

\section{Main Experiment Details}
\label{app:training-details}

For the main experiments in~\autoref{sec:experiments}, we first summarize the baseline-specific settings in~\autoref{tab:appendix-main-baseline-settings}.
The shared model configurations and training hyperparameters for the 1.5B and 3B Qwen3-style models are provided in~\autoref{tab:appendix-main-setup}.
For each model scale, all methods use the same base architecture and training setup, differing only in the corresponding cross-layer mechanism.

\begin{table}[H]
\centering
\scriptsize
\setlength{\tabcolsep}{2pt}
\renewcommand{\arraystretch}{1.12}
\resizebox{\textwidth}{!}{%
\begin{tabular}{@{}lccccc@{}}
\toprule
\textbf{Setting}
& \textbf{Vanilla}
& \textbf{Depth-Attention}
& \textbf{DenseFormer}
& \shortstack{\textbf{Residual}\\\textbf{Attention}}
& \textbf{mHC} \\
\midrule
Cross-layer mechanism
& -- 
& \shortstack{Depth-wise\\value mixing}
& \shortstack{Dense hidden-state\\reuse}
& \shortstack{Block-wise residual\\aggregation}
& \shortstack{Residual-stream\\mixing} \\
Main hyperparameter
& -- 
& Stride $s=24$
& \shortstack{Depth-weighted averaging}
& Block size 12
& 4 streams \\
Base architecture
& \multicolumn{5}{c}{Same as Table~\ref{tab:appendix-main-setup}} \\
Training setup
& \multicolumn{5}{c}{Same data, token budget, optimizer, schedule, and hyperparameters} \\
\bottomrule
\end{tabular}%
}
\caption{
Baseline-specific settings used in the main experiments.
All methods share the same base model configurations and training hyperparameters in~\autoref{tab:appendix-main-setup}.
mHC denotes manifold hyperconnection.
}
\label{tab:appendix-main-baseline-settings}
\end{table}

We choose the baseline-specific hyperparameters by following the recommended
or commonly used practical configurations in the corresponding papers, while
keeping the base architecture, data, optimizer, training budget, sequence length,
and precision fixed across methods. For DenseFormer~\cite{pagliardini2024denseformer}, we use the full
$1{\times}1$ setting, where a DWA module is applied after every Transformer block
and aggregates the current block output, all previous block outputs, and the
input embedding. For Attention Residuals~\cite{team2026attention}, we use the practical Block AttnRes variant with
eight stored block representations. In the Attention Residuals implementation,
the block size counts attention and MLP residual sublayers; since each
Transformer layer contains two such residual sublayers, our 48-layer model with
block size $12$ gives $2L/12=8$ blocks, matching the recommended practical
setting. For mHC~\cite{xie2025mhc}, we use the expansion rate $n=4$, following the standard mHC
configuration. These choices give each baseline a strong practical configuration
from its original design, rather than weakening the baselines through additional
sparsification or smaller cross-layer state.

\begingroup
\scriptsize
\setlength{\LTleft}{0pt}
\setlength{\LTright}{0pt}
\setlength{\LTcapwidth}{\textwidth}
\setlength{\tabcolsep}{4pt}
\renewcommand{\arraystretch}{0.95}
\captionof{table}{
Model configurations and training hyperparameters for the main 1.5B and 3B Qwen3-style experiments.
Depth-Attention-specific entries apply only to our method; the vanilla Transformer and cross-layer baselines use the same base architecture with their corresponding mechanism enabled.
}
\label{tab:appendix-main-setup}
\addtocounter{table}{-1}
\renewcommand{\theHtable}{\thetable.lt}
\begin{longtable}{@{}>{\raggedright\arraybackslash}p{0.48\textwidth}>{\centering\arraybackslash}p{0.22\textwidth}>{\centering\arraybackslash}p{0.22\textwidth}@{}}
\toprule
\textbf{Configuration / Hyperparameter} & \textbf{1.5B} & \textbf{3B} \\
\midrule
\endfirsthead
\multicolumn{3}{@{}l}{\textbf{Table~\ref{tab:appendix-main-setup} continued}} \\
\toprule
\textbf{Configuration / Hyperparameter} & \textbf{1.5B} & \textbf{3B} \\
\midrule
\endhead
\bottomrule
\endfoot
\multicolumn{3}{@{}l}{\textit{Model configuration}} \\
\midrule
Architecture & Qwen3-style decoder & Qwen3-style decoder \\
Hidden size & 1536 & 2048 \\
Intermediate size & 4096 & 6912 \\
Number of hidden layers & 48 & 48 \\
Number of attention heads & 24 & 32 \\
Number of key-value heads & 6 & 8 \\
GQA group size & 4 & 4 \\
Activation function & SiLU & SiLU \\
Vocabulary size & 50304 & 50304 \\
Maximum sequence length & 2048 & 2048 \\
Maximum position embeddings & 2048 & 2048 \\
RMSNorm epsilon & $1.0\times10^{-5}$ & $1.0\times10^{-5}$ \\
Initializer range & 0.01 & 0.01 \\
Tie word embeddings & False & False \\
BOS / EOS token id & 0 / 0 & 0 / 0 \\
Padding token id & 1 & 1 \\
Model dtype & FP16 & FP16 \\
Depth-Attention stride & 24 & 24 \\
\midrule
\multicolumn{3}{@{}l}{\textit{Training hyperparameters}} \\
\midrule
Training data & The Pile & The Pile \\
Training tokens & 32B & 32B \\
Training steps & 15625 & 15625 \\
Global batch size & 1024 & 1024 \\
Sequence length & 2048 & 2048 \\
Optimizer & AdamW & AdamW \\
Learning rate & $1.0\times10^{-3}$ & $1.0\times10^{-3}$ \\
Learning rate schedule & cosine decay & cosine decay \\
Warmup ratio & 0.02 & 0.02 \\
Adam $\beta_1$ & 0.9 & 0.9 \\
Adam $\beta_2$ & 0.95 & 0.95 \\
Weight decay & 0.1 & 0.1 \\
Gradient clipping & 1.0 & 1.0 \\
Precision & FP16 & FP16 \\
\end{longtable}
\endgroup

\section{Overhead Derivations and Benchmark Details}
\label{app:overhead-details}

This appendix provides the derivations for the theoretical overhead estimates in
\autoref{tab:theoretical-overhead} and the empirical efficiency measurements in
\autoref{tab:efficiency}. All theoretical estimates report the additional cost
beyond a vanilla Transformer decoder.

\subsection{Common Counting Assumptions}

We use the following notation. Let $L$ be the number of Transformer layers,
$d_{\mathrm{model}}$ be the hidden dimension, and $d_{\mathrm{kv}}$ be the total
key-value dimension used by the grouped-query attention cache. For our 3B
Qwen3-style configuration, we use
\[
L=48,\qquad d_{\mathrm{model}}=2048,\qquad d_{\mathrm{kv}}=512.
\]
For long-context memory estimates, we use a prefill length of
\[
T=128\mathrm{K}=131072
\]
tokens. All memory and activation-I/O estimates assume 16-bit activations,
so each scalar takes
\[
b=2
\]
bytes. This is consistent with the FP16 precision used in our experiments.

We count one multiply-add as two FLOPs. We focus on the dominant vector dot
products and weighted sums, and omit lower-order scalar operations such as the
softmax normalization over a small number of depth sources. For the normalized
FLOPs overhead in \autoref{tab:efficiency}, we follow the standard scaling-law
compute approximation that dense language-model training costs about $6ND$
FLOPs for $N$ parameters and $D$ tokens~\citep{kaplan2020scaling}. Since training
consists of a forward pass and a backward pass, this corresponds to a dense
forward cost of about $2N$ FLOPs per token. Therefore, for a 3B model, the
standard dense one-token forward/decode estimate is
\[
2N = 2 \times 3\times 10^9 = 6\times 10^9.
\]

\subsection{Depth-Attention}

For a target layer $\ell$, Depth-Attention mixes value states from a depth source
set $\mathcal{D}_{\ell}$. Let
\[
M_{\ell}=|\mathcal{D}_{\ell}|
\]
denote the number of depth sources at layer $\ell$. In our implementation, with
zero-indexed layer ids, the depth source set is maintained as
$\mathcal{D}_{\ell}=\{\ell\}\cup\{\, ms \mid m\ge 0,\ ms<\ell\,\}$ with
stride $s=24$, giving the layer-wise asymptotic cost
\[
O\!\left((\lceil \ell/s\rceil+1)d_{\mathrm{kv}}\right).
\]

At each target layer, Depth-Attention performs two dominant operations along the
depth dimension. First, it computes depth-wise scores between the current query
and the source keys:
\[
q_{\ell}K_{\mathcal{D}_{\ell}}^{\top},
\]
which costs approximately
\[
2M_{\ell}d_{\mathrm{kv}}
\]
FLOPs. Second, it computes the weighted sum over source values:
\[
\alpha_{\ell}V_{\mathcal{D}_{\ell}},
\]
which also costs approximately
\[
2M_{\ell}d_{\mathrm{kv}}
\]
FLOPs. Therefore, the extra FLOPs at layer $\ell$ are
\[
C_{\ell}^{\mathrm{Depth}}
\approx
4M_{\ell}d_{\mathrm{kv}}.
\]
Using the exact depth source schedule of our 3B model, we have
\[
\sum_{\ell=0}^{L-1}M_{\ell}=118.
\]
Thus, the total extra FLOPs are
\begin{align}
C^{\mathrm{Depth}}
&=
4d_{\mathrm{kv}}\sum_{\ell=0}^{L-1}M_{\ell} \notag\\
&=
4\times512\times118 \notag\\
&=
2.42\times 10^{5},
\end{align}
which gives $2.42\times 10^{5}$ extra FLOPs per token.

For activation I/O, Depth-Attention reads the source keys and source values and
writes the mixed value state. Thus the extra I/O at layer $\ell$ is counted as
\[
B_{\ell}^{\mathrm{Depth}}
\approx
b\,d_{\mathrm{kv}}(2M_{\ell}+1).
\]
Summing over all layers gives
\begin{align}
B^{\mathrm{Depth}}
&=
b\,d_{\mathrm{kv}}
\left(2\sum_{\ell=0}^{L-1}M_{\ell}+L\right) \notag\\
&=
2\times512\times(2\times118+48) \notag\\
&=
290816\ \mathrm{bytes} \notag\\
&\approx
284~\mathrm{KiB}.
\end{align}

Depth-Attention does not require storing additional hidden states during inference.
The mixed value state replaces the standard value state in the ordinary
key-value cache, and the current-layer key is unchanged. Therefore, the extra
persistent memory for a 128K-token prefill is
\[
0~\mathrm{GiB}.
\]

\subsection{DenseFormer}

DenseFormer-like dense cross-layer mechanisms aggregate hidden states from many
previous layers. For a target layer $\ell$, we count the number of accessible
hidden states as approximately $\ell+1$, giving the layer-wise cost
\[
O(\ell d_{\mathrm{model}}).
\]

The main computation is the weighted aggregation of $d_{\mathrm{model}}$-
dimensional hidden states. The per-layer FLOPs are
\[
C_{\ell}^{\mathrm{Dense}}
\approx
2(\ell+1)d_{\mathrm{model}}.
\]
We define
\[
S_{\mathrm{Dense}}
=
\sum_{\ell=1}^{L}(\ell+1)
=
1224 .
\]
The total extra FLOPs are then
\begin{align}
C^{\mathrm{Dense}}
&=
2d_{\mathrm{model}}S_{\mathrm{Dense}} \notag\\
&=
2\times 2048\times 1224 \notag\\
&=
5.01\times 10^{6},
\end{align}
which gives $5.01\times 10^{6}$ extra FLOPs per token.

For activation I/O, DenseFormer reads the previous hidden states and writes the
aggregated hidden state. Thus,
\begin{align}
B^{\mathrm{Dense}}
&=
b d_{\mathrm{model}}(S_{\mathrm{Dense}}+L) \notag\\
&=
2\times 2048\times(1224+48) \notag\\
&=
5.21\times 10^{6}\ \mathrm{bytes} \notag\\
&\approx
4.97\ \mathrm{MiB}.
\end{align}

For long-context prefill, DenseFormer must keep the hidden states of all layers.
The extra persistent memory is therefore
\[
M^{\mathrm{Dense}}
=
T L d_{\mathrm{model}} b.
\]
Plugging in $T=131072$, $L=48$, $d_{\mathrm{model}}=2048$, and $b=2$, we obtain
\begin{align}
M^{\mathrm{Dense}}
&=
131072\times 48\times 2048\times 2\ \mathrm{bytes} \notag\\
&=
24.0\ \mathrm{GiB}.
\end{align}

\subsection{Attention Residuals}

Attention Residuals attends over a set of residual blocks at each target layer.
Let $R_{\ell}$ denote the number of residual blocks used at layer $\ell$. Its
layer-wise cost is therefore
\[
O(R_{\ell}d_{\mathrm{model}}).
\]

The dominant operations are residual-block scoring and the weighted sum over
residual representations. We count these as
\[
C_{\ell}^{\mathrm{ResAttn}}
\approx
4R_{\ell}d_{\mathrm{model}}.
\]
We denote the total number of residual blocks used across all target layers by
\[
S_{\mathrm{Res}}
=
\sum_{\ell=1}^{L}R_{\ell}
=
520 .
\]
The total extra FLOPs are
\begin{align}
C^{\mathrm{ResAttn}}
&=
4d_{\mathrm{model}}S_{\mathrm{Res}} \notag\\
&=
4\times 2048\times 520 \notag\\
&=
4.26\times 10^{6},
\end{align}
which gives $4.26\times 10^{6}$ extra FLOPs per token.

The extra activation I/O is computed by counting the residual-block states read
by the attention module and the additional residual-attention outputs written by
the module. After subtracting the standard vanilla residual I/O, we obtain
\begin{align}
B^{\mathrm{ResAttn}}
&\approx
b d_{\mathrm{model}}
\left(2S_{\mathrm{Res}}+2L\right) \notag\\
&=
2\times 2048\times(2\times520+2\times48) \notag\\
&=
4.65\times 10^{6}\ \mathrm{bytes} \notag\\
&\approx
4.44\ \mathrm{MiB}.
\end{align}

For the 128K-prefill memory estimate, Attention Residuals stores a fixed set of
residual-block hidden states. In our comparison, the number of stored residual
blocks is $R=8$, so the extra persistent memory is
\[
M^{\mathrm{ResAttn}}
=
T R d_{\mathrm{model}} b.
\]
Plugging in $T=131072$, $R=8$, $d_{\mathrm{model}}=2048$, and $b=2$, we obtain
\begin{align}
M^{\mathrm{ResAttn}}
&=
131072\times 8\times 2048\times 2\ \mathrm{bytes} \notag\\
&=
4.0\ \mathrm{GiB}.
\end{align}

\subsection{mHC}

For mHC, let $n$ denote the total number of residual streams.
Following the standard mHC configuration, we use an expansion rate of $n=4$,
i.e., the model maintains four residual streams in total.
The method introduces interactions among these streams, giving a layer-wise
stream-mixing computation cost of
\[
O(n^3 d_{\mathrm{model}})
\]
and an activation I/O cost of
\[
O(n d_{\mathrm{model}}).
\]

Using the operation count of the mHC stream-mixing module under this
configuration, the extra FLOPs over the full model are
\[
C^{\mathrm{mHC}}
\approx
4.72\times 10^{7}
\]
extra FLOPs per token. The corresponding activation I/O count gives
\[
B^{\mathrm{mHC}}
\approx
8.63~\mathrm{MiB}
\]
extra I/O per token.

For long-context prefill, mHC needs to keep multiple residual streams.
Since a vanilla Transformer already has one residual stream, the extra
persistent memory over vanilla comes from the additional $n-1$ streams:
\[
M^{\mathrm{mHC}}
=
T (n-1) d_{\mathrm{model}} b.
\]
With $T=131072$, $n=4$, $d_{\mathrm{model}}=2048$, and $b=2$,
\[
M^{\mathrm{mHC}}
=
131072\times (4-1)\times 2048\times 2
=
1.5~\mathrm{GiB}.
\]
Thus, although mHC maintains four residual streams in total, its reported
extra memory over a vanilla Transformer corresponds to the three additional
streams.

\subsection{Empirical Benchmark Details}

We further measure the practical efficiency of the trained 3B Qwen3-style model.
All empirical numbers in \autoref{tab:efficiency} are measured on trained
checkpoints under the same hardware and software environment.

For training speed, we report the average wall-clock time per optimization step.
The vanilla model takes
\[
4.63\ \mathrm{s/step},
\]
while Depth-Attention takes
\[
5.15\ \mathrm{s/step}.
\]
The relative training overhead is therefore
\[
\frac{5.15-4.63}{4.63}\times 100\%
=
11.23\%.
\]

For inference, we use batch size $64$, prefill length $2048$, and decode length
$2048$. We use one warmup run and two repeated measurement runs. The table
reports the total wall-clock generation time, including both prefill and
decoding. The vanilla model takes
\[
526.6\ \mathrm{s},
\]
while Depth-Attention takes
\[
532.8\ \mathrm{s}.
\]
The relative inference overhead is therefore
\[
\frac{532.8-526.6}{526.6}\times 100\%
=
1.18\%.
\]

The prefill peak memory is
\[
30.6\ \mathrm{GB}
\]
for both vanilla and Depth-Attention, so the reported prefill-memory overhead is
\[
0.0\ \mathrm{GB}.
\]

Finally, the extra FLOPs entry in \autoref{tab:efficiency} uses the theoretical
Depth-Attention estimate derived above:
\[
2.42\times 10^{5}
\]
extra FLOPs per token. The overhead row normalizes this value by the standard
dense one-token forward estimate $2N$ for a 3B model:
\[
\frac{2.42\times 10^{5}}{2\times 3\times 10^{9}}
=
4.0\times 10^{-5}
=
0.004\%.
\]
This explains why the theoretical arithmetic overhead is negligible even though
the measured training wall-clock overhead is larger: the practical overhead is
mainly caused by additional activation reads/writes, kernel launches, and
non-fused depth-wise operations rather than by dense matrix-multiplication FLOPs.

\section{Scaling Experiment Details}
\label{app:scaling-details}

For the scaling experiments in~\autoref{sec:scaling}, we provide the model configurations and training hyperparameters used for the Qwen-style scaling analysis in~\autoref{tab:appendix-qwen-scaling-setup}.
For each model size, Depth-Attention and the vanilla Transformer use the same base configuration and optimization setup, differing only in whether depth-wise value mixing is enabled.
The Qwen-style scaling experiments include 360M, 500M, and 710M models.
The earlier 160M setting is not included in the final scaling analysis.
The larger 1.5B and 3B Qwen3-style models follow the configurations and training setup in~\autoref{tab:appendix-main-setup}.

\begingroup
\small
\setlength{\LTleft}{0pt}
\setlength{\LTright}{0pt}
\setlength{\LTcapwidth}{\textwidth}
\setlength{\tabcolsep}{4pt}
\renewcommand{\arraystretch}{0.98}
\captionof{table}{
Model configurations and training hyperparameters for the Qwen-style scaling experiments.
Depth-Attention-specific entries apply only to our method; vanilla baselines use the same base configurations with depth-wise value mixing disabled.
The 1.5B and 3B scaling points use the main-experiment setup in Table~\ref{tab:appendix-main-setup}.
}
\label{tab:appendix-qwen-scaling-setup}
\addtocounter{table}{-1}
\renewcommand{\theHtable}{\thetable.lt}
\begin{longtable}{@{}>{\raggedright\arraybackslash}p{0.34\textwidth}>{\centering\arraybackslash}p{0.19\textwidth}>{\centering\arraybackslash}p{0.19\textwidth}>{\centering\arraybackslash}p{0.19\textwidth}@{}}
\toprule
\textbf{Configuration / Hyperparameter}
& \textbf{360M}
& \textbf{500M}
& \textbf{710M} \\
\midrule
\endfirsthead
\multicolumn{4}{@{}l}{\textbf{Table~\ref{tab:appendix-qwen-scaling-setup} continued}} \\
\toprule
\textbf{Configuration / Hyperparameter}
& \textbf{360M}
& \textbf{500M}
& \textbf{710M} \\
\midrule
\endhead
\bottomrule
\endfoot
\multicolumn{4}{l}{\textit{Model configuration}} \\
\midrule
Architecture & Qwen-style decoder & Qwen-style decoder & Qwen-style decoder \\
Hidden size & 960 & 1152 & 1280 \\
Intermediate size & 2540 & 3072 & 3072 \\
Number of hidden layers & 24 & 24 & 32 \\
Number of attention heads & 15 & 18 & 20 \\
Number of key-value heads & 15 & 18 & 20 \\
GQA group size & 1 & 1 & 1 \\
Activation function & SiLU & SiLU & SiLU \\
Vocabulary size & 50304 & 50304 & 50304 \\
Maximum sequence length & 2048 & 2048 & 2048 \\
Maximum position embeddings & 2048 & 2048 & 2048 \\
RMSNorm epsilon & $1.0\times10^{-5}$ & $1.0\times10^{-5}$ & $1.0\times10^{-5}$ \\
Initializer range & 0.02 & 0.02 & 0.02 \\
Tie word embeddings & False & False & False \\
BOS / EOS token id & 0 / 0 & 0 / 0 & 0 / 0 \\
Padding token id & 1 & 1 & 1 \\
Model dtype & FP16 & FP16 & FP16 \\
Depth-Attention stride & 12 & 12 & 16 \\
\midrule
\multicolumn{4}{l}{\textit{Training hyperparameters}} \\
\midrule
Training data & The Pile & The Pile & The Pile \\
Training tokens & 32B & 32B & 32B \\
Training steps & 15625 & 15625 & 15625 \\
Global batch size & 1024 & 1024 & 1024 \\
Sequence length & 2048 & 2048 & 2048 \\
Optimizer & AdamW & AdamW & AdamW \\
Learning rate & $1.5\times10^{-3}$ & $1.0\times10^{-3}$ & $1.0\times10^{-3}$ \\
Learning rate schedule & cosine decay & cosine decay & cosine decay \\
Warmup ratio & 0.02 & 0.02 & 0.02 \\
Adam $\beta_1$ & 0.9 & 0.9 & 0.9 \\
Adam $\beta_2$ & 0.95 & 0.95 & 0.95 \\
Weight decay & 0.1 & 0.1 & 0.1 \\
Gradient clipping & 1.0 & 1.0 & 1.0 \\
Precision & FP16 & FP16 & FP16 \\
\end{longtable}
\endgroup
\end{document}